\documentclass[10pt,twocolumn,letterpaper]{article}

\usepackage{cvpr}              % To produce the CAMERA-READY version
\usepackage{graphicx}
\usepackage{amsmath}
\usepackage{amssymb}
\usepackage{booktabs}
\usepackage{multirow}
\usepackage{amssymb}
\usepackage{array}
\usepackage{bbm}
\usepackage{marvosym}
\usepackage[accsupp]{axessibility}  % Improves PDF readability for those with disabilities.

\newcolumntype{L}[1]{>{\raggedright\let\newline\\\arraybackslash\hspace{0pt}}m{#1}}
\newcolumntype{C}[1]{>{\centering\let\newline\\\arraybackslash\hspace{0pt}}m{#1}}
\newcolumntype{R}[1]{>{\raggedleft\let\newline\\\arraybackslash\hspace{0pt}}m{#1}}

\usepackage[pagebackref,breaklinks,colorlinks]{hyperref}

\usepackage[capitalize]{cleveref}
\crefname{section}{Sec.}{Secs.}
\Crefname{section}{Section}{Sections}
\Crefname{table}{Table}{Tables}
\crefname{table}{Tab.}{Tabs.}

\newcommand{\squishlist}{
 \begin{list}{$\bullet$}
  { \setlength{\itemsep}{0pt}
     \setlength{\parsep}{1pt}
     \setlength{\topsep}{1pt}
     \setlength{\partopsep}{0pt}
     \setlength{\leftmargin}{1.5em}
     \setlength{\labelwidth}{1em}
     \setlength{\labelsep}{0.5em} } }
\newcommand{\squishend}{
  \end{list}  }

\def\cvprPaperID{9804} % *** Enter the CVPR Paper ID here
\def\confName{CVPR}
\def\confYear{2022}

\begin{document}

\title{Class-Aware Contrastive Semi-Supervised Learning}

\renewcommand{\thempfootnote}{\arabic{mpfootnote}}
\author{Fan Yang$^2$\thanks{Equal contribution.} \hspace{0.3cm} 
Kai Wu$^1$\footnotemark[\value{footnote}] \hspace{0.3cm} 
Shuyi Zhang$^1$ \hspace{0.3cm} 
Guannan Jiang$^4$ \hspace{0.3cm}
Yong Liu$^1$ \hspace{0.3cm} \\
Feng Zheng$^3$\thanks{Corresponding author.} \hspace{0.3cm}
Wei Zhang$^4$ \hspace{0.3cm}
Chengjie Wang$^1$\footnotemark[\value{footnote}] \hspace{0.3cm}
Long Zeng$^2$\\
% $^1$Tencent Youtu Lab, Shenzhen, China  \qquad $^2$Tsinghua University, Shenzhen, China \\
% $^3$Southern University of Science and Technolog, Shenzhen, China  \qquad $^4$CATL, Ningde, China\\
$^1$Tencent Youtu Lab \hspace{0.3cm} $^2$Tsinghua University \hspace{0.3cm} $^3$Southern University of Science and Technolog \hspace{0.3cm} $^4$CATL\\
{\tt\small fan-yang20@mails.tsinghua.edu.cn, \{lloydwu, shuyizhang, choasliu, jasoncjwang\}@tencent.com} \\
{\tt\small zfeng02@gmail.com, \{jianggn, zhangwei\}@catl.com, zenglong@sz.tsinghua.edu.cn}
}
\maketitle

%%%%%%%%% ABSTRACT
\begin{abstract}
Pseudo-label-based semi-supervised learning (SSL) has achieved great success on raw data utilization. However, its training procedure suffers from confirmation bias due to the noise contained in self-generated artificial labels. Moreover, the model's judgment becomes noisier in real-world applications with extensive out-of-distribution data. To address this issue, we propose a general method named Class-aware Contrastive Semi-Supervised Learning (CCSSL), which is a drop-in helper to improve the pseudo-label quality and enhance the model's robustness in the real-world setting. Rather than treating real-world data as a union set, our method separately handles reliable in-distribution data with class-wise clustering for blending into downstream tasks and noisy out-of-distribution data with image-wise contrastive for better generalization. Furthermore, by applying target re-weighting, we successfully emphasize clean label learning and simultaneously reduce noisy label learning. Despite its simplicity, our proposed CCSSL has significant performance improvements over the state-of-the-art SSL methods on the standard datasets CIFAR100~\cite{krizhevsky2009learning} and STL10~\cite{coates2011analysis}. On the real-world dataset Semi-iNat 2021~\cite{su2021semi_iNat}, we improve FixMatch~\cite{sohn2020fixmatch} by 9.80\% and CoMatch~\cite{li2021comatch} by 3.18\%. Code is available  https://github.com/TencentYoutuResearch/Classification-SemiCLS.

\end{abstract}

%%%%%%%%% BODY TEXT
\section{Introduction}

\begin{figure}
\centering
\begin{minipage}[b]{1.0\linewidth}
    \centering
    \subfloat[][Real-World Data With In-Distribution and Out-of-Distribution Data]{\label{fig1_a}\includegraphics[width=1.0\linewidth]{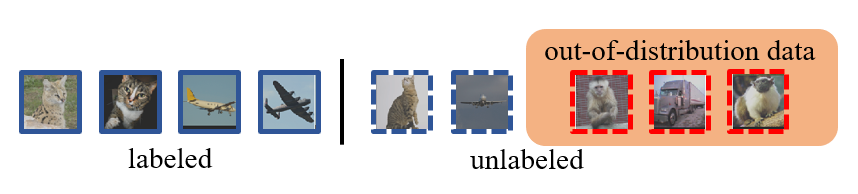}}
\end{minipage} %\par

\medskip
\begin{minipage}[b]{1.0\linewidth}
    \centering
    \subfloat[justification=centering][Pseudo-Label-Based SSL]{\label{fig1_b}\includegraphics[width=0.5\linewidth]{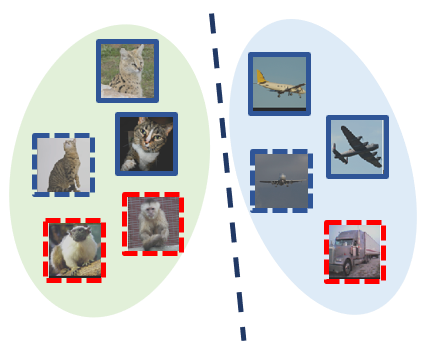}} 
    \subfloat[justification=centering][Class-Aware Contrastive SSL (ours)]{\label{fig1_c}\includegraphics[width=0.5\linewidth]{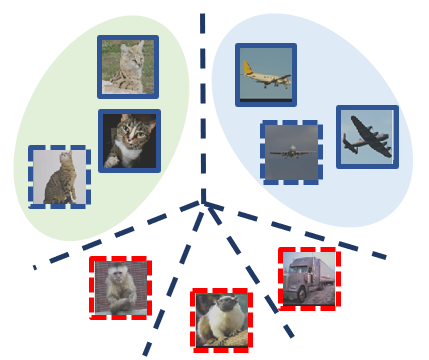}} 
\end{minipage}

\caption{Intuition graph for class-aware contrastive semi-supervised learning.~\subref{fig1_a} represents real-world unlabeled data containing in-distribution and out-of-distribution data (unbalanced distribution or unknown classes). Unlike pseudo-label-based semi-supervised learning in~\subref{fig1_b}, which wrongly clusters noisy out-of-distribution data, CCSSL in~\subref{fig1_c} reduces noise by image-level contrastive learning on out-of-distribution data while maintaining the class-aware clustering ability for in-distribution data.}
%  reduces noises by contrastive learning while maintaining the class-aware clustering ability for downstream SSL tasks. 
\label{fig:intuitive}
\end{figure}

\label{sec:intro}
Raw data utilization is becoming a research focal point for its high accessibility and high affordability. The definition of raw data is the union set of in-distribution data (known classes and balanced distribution) and out-of-distribution data~\cite{xu2021stable} (unknown classes or unbalanced distribution), as shown in \cref{fig:intuitive}. For in-distribution datasets, semi-supervised learning (SSL) has achieved excellent performance with the help of pseudo labeling~\cite{sohn2020fixmatch}~\cite{berthelot2019mixmatch}~\cite{berthelot2019remixmatch}~\cite{hu2021simple}. The primary process for the pseudo-label based SSL is iterative training: 1) creating self-generated pseudo labels on raw data, 2) training on pseudo labels, 3) repeating 1 and 2. The underlying assumption for training on pseudo labels is that the distribution of the labeled data is close to the unlabeled, and the unlabeled dataset does not contain any novel categories. This assumption often does not hold in the real-world applications with extensive out-of-distribution data, which contains unbalanced distribution or unknown classes. SSL's training and labeling loop collapses in the real-world because of the enormous noise introduced by pseudo labeling on out-of-distribution data. The confirmation bias~\cite{arazo2020pseudo} induced by noisy pseudo labels deteriorates SSL performance by a large margin~\cite{su2021realistic}.

SSL has used many techniques to alleviate the confirmation bias and achieved state-of-the-art results on standard in-distribution benchmarks, like CIFAR~\cite{krizhevsky2009learning} and STL10~\cite{coates2011analysis}. Some methods achieve this goal by using the model's self-correcting capability. In~\cite{sohn2020fixmatch}~\cite{berthelot2019mixmatch}~\cite{berthelot2019remixmatch}~\cite{sohn2020fixmatch}, a high confidence threshold is used to filter incorrectly pseudo-labeled data. ~\cite{zhou2021instant}~\cite{liu2021unbiased}~\cite{tang2021humble} use a self-ema teacher to generate pseudo labels with the assumption that the weighted averaged model produces more stable and less noisy predictions. Some methods~\cite{taherkhani2021self}~\cite{wang2021data} try to narrow the distribution gap between raw data and labeled data by predictions' uncertainty. However, justifying a model's prediction by its output without introducing other information still suffers from confirmation bias~\cite{nassar2021labels}. Especially on the real-world data with unbalanced distribution or unknown classes, the effect of a model's self-correcting is facing a huge challenge. By explicitly introducing contrastive information, our CCSSL alleviates the confirmation bias in the feature space and shows great de-noise ability on the real-world data.

Another trend is using prior or posterior information to co-rectify a model's predictions. Both~\cite{he2020momentum}~\cite{chen2020simple} use a pre-trained model as a universal prior information. In~\cite{nassar2021labels}, by knowledge embedding graph, semantic information is used to regularize feature learning. Although the above methods are proved helpful, fixed prior is hard to be blended into downstream tasks, resulting in inferior performance in practice.~\cite{li2021comatch} utilizes posterior information by constructing a prediction graph and introducing contrastive learning from self-supervised learning (Self-SL). However, directly combining image-level feature repulsion interferes with SSL's class-clustering ability. To solve this conflict, in our proposed CCSSL, a class-aware contrastive module is specifically explored for reliable in-distribution data clustering and noisy out-of-distribution data contrasting to better integrate with downstream classification tasks. 

To this end, we propose a class-aware contrastive semi-supervised learning (CCSSL) method. CCSSL consists of a semi-supervised module and a class-aware contrastive module. Any end-to-end pseudo-label-based SSL can replace the semi-supervised module to benefit from the class-aware contrastive module's confirmation bias alleviation ability. Unlike self-correcting methods that try to alleviate noise by model's own output, CCSSL regularizes the training process by introducing information in the feature space. As shown in the \cref{fig:intuitive}, CCSSL constructs the feature space with high dimensional vectors from two strong augmented views. Rather than directly combining contrastive learning, CCSSL uses class-aware clustering on in-distribution data to maintain SSL's clustering ability and image-level contrasting on out-of-distribution data for noise alleviation. Furthermore, to reduce the implicit noise introduced from the class-aware clustering, we incorporate a target re-weighting module that emphasizes learning on high probable in-distribution samples and reduces the effect of uncertain noisy samples. We found that with the help of the re-weighting module, the prediction confidence of learned models is robust enough for roughly discriminating in-distribution and out-of-distribution data. 

Our contributions can be described in three folds:
\squishlist
\item We propose a novel SSL learning method CCSSL, which takes advantage of both SSL's effective learning and Self-SL's noise alleviation ability by class-wise clustering on in-distribution data and image-wise contrasting on out-of-distribution data.  

\item CCSSL is a drop-in helper that can improve upon any end-to-end pseudo-label-based SSL method for confirmation bias alleviation and faster convergence. CCSSL's noise alleviation capability makes SSL methods more practical in the real-world setting.  

\item We thoroughly analyzed the effect of CCSSL with various state-of-the-art SSL methods on both in-distribution data and real-world data. By simply combining CCSSL, current state-of-the-art SSL methods can be further improved.
\squishend

\begin{figure*}
    \centering
    \small
      \includegraphics[width=\textwidth, height=\linewidth, keepaspectratio]{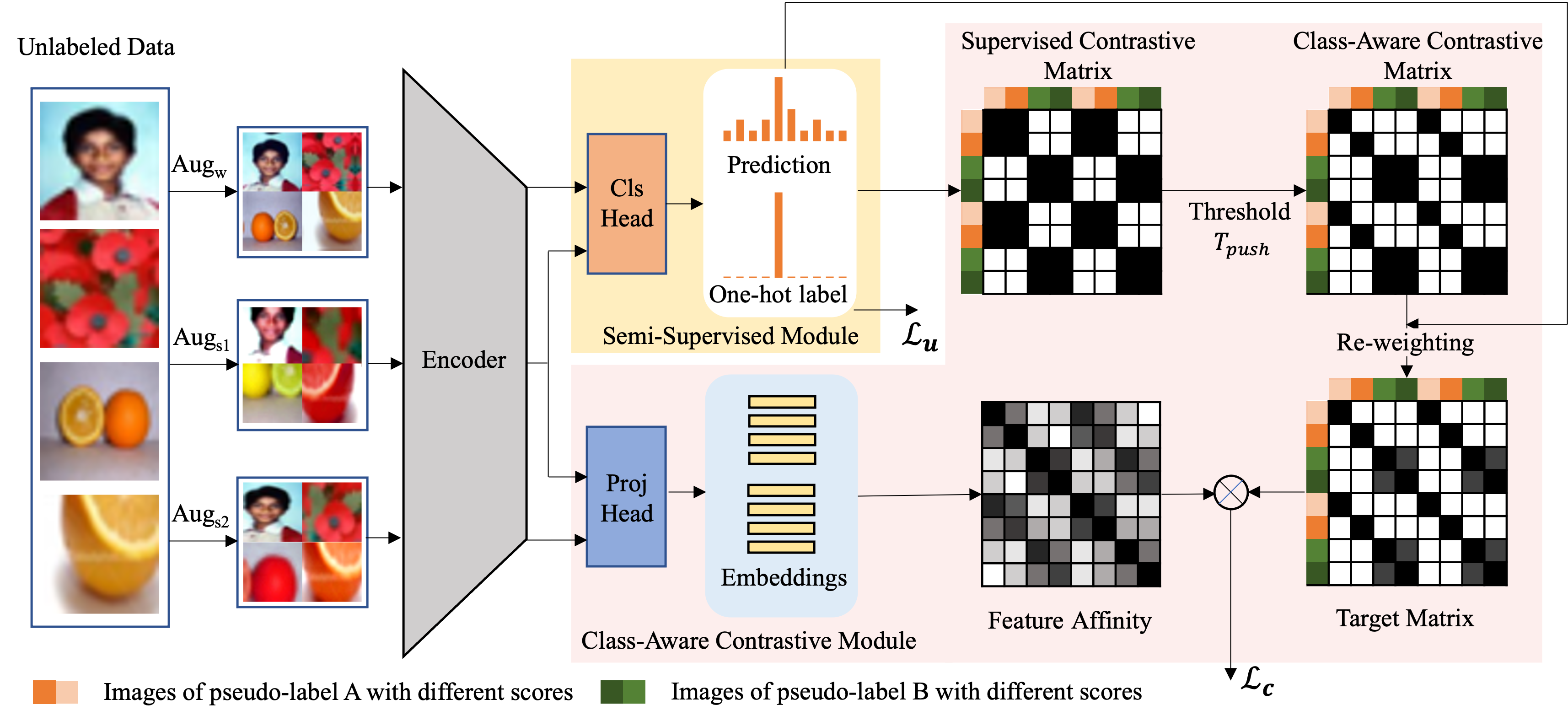}
    \caption{Framework of our proposed CCSSL. Given a batch of unlabeled images, the weakly augmented views will go through a semi-supervised module that can be borrowed from any pseudo-label-based SSL to generate model predictions. With pseudo labels, we make a supervised contrastive matrix with only class-level information. Then, the class-aware contrastive matrix is formed by image-level contrasting on out-of-distribution data to reduce confirmation bias. By applying a re-weighting module, we focus learning on clean data and get the final target matrix. Besides, feature affinity matrix is made by two strong augmented views. Class-aware contrastive module for unlabeled data is formulated by minimizing cross-entropy between the affinity matrix and the target matrix.}
    \label{fig:method}
\end{figure*}

%------------------------------------------------------------------------
\section{Related Work}
\label{sec:related}
In this section, we describe the recent trends of deep learning based SSL and contrastive based Self-SL. Then we analyze specific methods aiming at using on the real-world setting. 

\subsection{Semi-Supervised Learning}
Deep learning based SSL mainly consists of pseudo-labeling and consistency regularization. Pseudo-label-based methods utilize unlabeled data by training on self-generated predictions. The self-training process validates the model itself and can also be named as self-confirming~\cite{dictionary2002merriam}. However, using pseudo labels suffers confirmation bias~\cite{arazo2020pseudo} because it is easy to overfit incorrect predictions during training. Both ~\cite{sohn2020fixmatch}~\cite{berthelot2019mixmatch} use high confidence predictions to filter noisy data. ~\cite{iscen2019label} makes a detour by label propagation to spread labeled data distribution on unlabeled data. Consistency based methods~\cite{jeong2019consistency}~\cite{tang2021humble}~\cite{sohn2020fixmatch} aim to produce consistent predictions on different image views based on the manifold assumption that different views of the same image should lie on the same point in the high dimensional space. ~\cite{sohn2020fixmatch}~\cite{tarvainen2017mean} create different views of an image by strong augmentations to simulate image perturbations of an object. Lately, pseudo-label-based consistency training~\cite{sohn2020fixmatch}~\cite{berthelot2019mixmatch}~\cite{berthelot2019remixmatch}, which use a weak augmentation for creating pseudo labels and a strong augmentation for consistency training, has dominated and achieved excellent performance gain. However, the research on the real-world setting with extensive out-of-distribution data has been less studied. With evaluating in-distribution data and real-world data, we show CCSSL's ability of alleviating confirmation bias in practical.  

\subsection{Self-Supervised Learning}
Self-SL has a long story for learning a universal representation without supervision. ~\cite{hinton2002training}~\cite{hyvarinen2005estimation} build the base for contrastive learning by proposing and refining the noise contrastive estimation (NCE). Based on NCE, ~\cite{oord2018representation} proposes infoNCE which inspires learning representations to discriminate different views of a sample among other samples. Moco~\cite{he2020momentum} and simCLR~\cite{chen2020simple} train a siamses network~\cite{chen2021exploring} to learn image-wise feature by pulling similar (positive) samples and pushing apart different (negative) samples~\cite{chen2021empirical}. Inspired by contrastive learning, we formulate positive pairs and negative pairs in CCSSL to regularize feature training. However, directly combining contrastive learning ignores the internal conflict between image-level feature repulsion and SSL's class-level feature clustering. By incorporating class-level information, CCSSL can seamlessly blend into downstream tasks.  

\subsection{Real-World Unlabeled Data Learning}
Real-world unlabeled data usually contains out-of-distribution data with unbalanced class distribution or unknown classes. Current pseudo-label-based SSL suffers from confirmation bias by incorrect pseudo-labeling on similar unknown classes or blurry images from out-of-distribution data, resulting in worse performance than supervised baseline~\cite{su2021realistic}. Open set classification is a subfield of real-world unlabeled data learning focusing on novel categories.~\cite{guo2020safe} keeps tracking the effect of the supervised learning model to prevent performance hazards. By incorporating deep-construction-based representation learning,~\cite{yoshihashi2019classification} learns to distinguish unknowns from knowns. Open set classification methods are proven to be effective on handwriting datasets or self-constructed datasets but seldom generalize on real-world data.~\cite{su2021realistic} conducts a systematic evaluation of several recently proposed SSL techniques on a real-world setting by two challenging datasets and shows that recent SSL's performance pales on real-world datasets in comparison to in-distribution standard datasets. Therefore, this paper further explores the real-world setting and narrows the gap between research and real-world applications by confirmation bias alleviation for out-of-distribution data. 
\section{Method}

\label{sec:method}
\subsection{Problem Definition}
In the setting of a semi-supervised classification task, let $\mathcal{X} = \{(x_i,y_i):i \in(1,...,B)\}$ be a batch of labeled data , where $x_i$ is the $i$th image and $y_i$ represents its one-hot label, and $\mathcal{U} = \{u_i:i \in(1,...,{\mu}B)\}$ be a batch of unlabeled data , where $\mu$ is a hyperparameter that determines the relative ratio of  $\mathcal{X}$ and $\mathcal{U}$. In the real world, unlabeled data $\mathcal{U}$ contains in-distribution data $\mathcal{U}_{in}$ and out-of-distribution data $\mathcal{U}_{out}$,  which makes pseudo labels more susceptible to confirmation bias~\cite{nassar2021labels}.

\subsection{Framework}
Our proposed CCSSL consists of a replaceable semi-supervised module and a class-aware contrastive module, as in ~\cref{fig:method}. Our method focuses on introducing class-aware contrastive information to alleviate confirmation bias for semi-supervised module especially in real-world setting. Any pseudo-label-based SSL can be used as the semi-supervised module to enjoy the benefit. The notations are as follows:

\textbf{Data augmentation} provides different views of a single image. For a labeled image $x_i$, a weak augmentation is used for supervised learning. For an unlabeled image $u_i$, we apply a weak augmentation $Aug_w(\cdot)$ and two strong augmentations $Aug_s(\cdot)$. The $Aug_w$ and one $Aug_s$ are sent to semi-supervise module, while all $Aug_s$ are used for class-aware contrastive learning.
        
\textbf{Encoder} $F(\cdot)$ is used to extract representation $r=F(Aug(x))$ for a given input $x$.
    % The resnet and the wideresnet series~\cite{sohn2020fixmatch} are used as encoders in our experiments, which maps the input into high-dimensional representation $r$ for downstream task
    
\textbf{Semi-Supervised module} can be replaced by any pseudo-label based semi-supervised learning method. The module generates pseudo labels during training and outputs the final prediction at inference time by a classification head $P_{cls}(\cdot)$. We use FixMatch~\cite{sohn2020fixmatch} in the method for simplicity.
    
\textbf{Class-Aware contrastive module} constructs an embedding space by mapping the high-dimensional representation $r$ to a low-dimensional embedding $z$ through the 
    projection head $P_{proj}(\cdot)$. We propose the class-aware contrastive learning in the embedding space to improve the robustness for out-of-distribution data and compatibleness with downstream tasks. 
    
\subsection{Semi-Supervised Module} \label{method:semi-sup-module}
Following the general form of pseudo-label-based SSL, the module contains a supervised loss $\mathcal{L}_{x}$ and an unsupervised loss $\mathcal{L}_{u}$. $\mathcal{L}_{x}$ is the same as~\cite{sohn2020fixmatch}. $\mathcal{L}_{u}$  relies on pseudo labels $\hat{q}_i = \arg\max(p_i)$ from the model's prediction $p_i = P_{cls}(Aug_w(u_i))$ on the weak view of image $u_i$. Only pseudo labels with high confidence $ q \geq t$, where $q = \max(p)$, will be retained for consistency training. H means cross entropy:

\begin{equation}
    \mathcal{L}_{u}=\frac{1}{\mu B} \sum_{i=1}^{\mu B}\mathbbm{1}(\max(p_i) \geq t)H(\hat{q_i}, P_{cls}(Aug_s(u_i)). \label{eq:pseudolabel}
\end{equation}

However, the manual set threshold $t$ is difficult to guarantee the accuracy of pseudo labels $\hat{q}$, especially on real-world setting with out-of-distribution data. The phenomenon of overfitting on self-generated labels is called confirmation bias~\cite{arazo2020pseudo}, and it can hardly be alleviated by the model itself during training. Therefore, we propose a novel class-aware contrastive learning method, which introduces unbiased information in the feature space. 

\subsection{Class-Aware Contrastive Module} \label{method:ClassAwareContrastive}

In \cref{fig:method}, we seamlessly integrate clustering and contrasting in the feature space and apply re-weighting for attentive training. Given unlabeled images in a batch, we separate images into two parts based on the likelihood of in-distribution and out-of-distribution, in which $T_{push}$ is used as the threshold for separation from the model's prediction confidence.  For highly possible in-distribution data $\max(p) > T_{push}$, we apply class-aware clustering in the feature space to seamlessly blend into downstream tasks. For out-of-distribution data $\max(p) < T_{push}$, we follow contrastive learning mechanism~\cite{chen2020simple} by taking a image's views as positive samples and others as negative. Then we apply foreground probability re-weighting for each sample to focus learning on high confidence clean data. Next, we will show how to formulate the method step by step. 

\textbf{Contrastive learning} aims to learn a universal prior information for downstream tasks. As in~\cite{he2020momentum}~\cite{chen2020simple}, we randomly sample a minibatch of $N$ images and each image $u_i$  has two strong augmentations $Aug_s$ which are used to extract features $r$ through encoder $F(\cdot)$. Then, $r$ is mapped to obtain a square normalized low-dimensional embedding $z$ by the projection head. Finally, we get an embedding's affinity matrix $S \in \mathbb{R}^{2N \times 2N}$ by the dot product of embeddings $\mathcal{Z}=\{z_i : i=1,...,2N\}$ with a temperature factor $\tau$, as follow:  
\begin{equation}
    s_{ij} = \mathrm{exp} (z_i \cdot z_j/\tau ) \label{eq:dotproduct},
\end{equation}
Then, a constrastive matrix $\mathbf{W}_{con} \in \mathbb{R}^{2N \times 2N}$ for loss calculation with $S$ is formulated as:
\begin{equation}
    w_{ij}^{con} = \left\{ \begin{array}{ll}  
    1 & \mathrm{if\ }j = i, \\
1 & \mathrm{if\ }z_i\ \mathrm{and\ } z_j\ \mathrm{are\ from\ the\ same\ image's}\\ & \mathrm{  views}, \\
0 & \mathrm{otherwise}. 
\end{array} \right. \label{eq:affinity}
\end{equation}

The self-supervised contrastive loss InfoNCE~\cite{oord2018representation} takes the following format:

\begin{equation}
    \mathcal{L}^{\text{InfoNCE}}=-\sum_{i=1}^{2N}\mathrm{log}\frac{\mathrm{exp}(z_i\cdot z_i^{*}/\tau) }{\sum_{j=1}^{2N}\mathbbm{1}_{(j\ne{i})} \mathrm{exp}(z_i\cdot z_{j}/\tau)},   
\end{equation}
% The methodology is to learn representations that attract similar (positive) samples and dispel different (negative) samples~\cite{chen2021empirical}. 
where $z_i^{*}$ is from the other $Aug_s$ of the same image as $z_i$. The ideal solution for contrastive learning is to push each image far from others and pull same image views closer in the feature space. The property is effective to alleviate noise on the out-of-distribution data. However, it is not compatible with the classification task which requires to cluster images at the class level. To solve the conflict between contrastive learning and downstream tasks, we formulate the class-aware contrastive module to take advantages of both Self-SL's generalizability and SSL's efficient training.

\textbf{Class-Aware Contrastive learning} considers class-aware information for in-distribution data while alleviating noise for out-of-distribution data. Inspired by~\cite{khosla2020supervised}, we consider the embeddings $z$ from the same category as positive pairs instead of the same image with the help of the pseudo-labels $\hat{q}$ from the semi-supervised module to form supervised contrastive matrix $W_{scon}$ in \cref{fig:method}.  However, class-wise clustering by pseudo-labels is unsafe to use because a model's prediction may contain a large amount of noises especially for out-of-distribution data, as explained in \cref{sec:intro}. Therefore, we introduce contrastive learning for noise regularization. We use $T_{push}$ as the threshold for clustering and contrasting in the feature space, and then $W_{scon}$ is tranformed to class-aware contrastive matrix $W_{clacon}$ as shown in \cref{fig:method}.  For model's prediction $p > T_{push}$, we assume the images have a high probability to be in-distribution data $\mathcal{U}_{in}$ and should be pulled closer with the same class. For $p < T_{push}$, our formula degenerates to contrastive learning where only $Aug_s$ of the same image are positive pairs. Each element $w_{ij}^{clacon}$ in $W_{clacon}$ has the following format: 
\begin{equation}
    w_{ij}^{clacon} = \left\{ \begin{array}{ll}  
1 & \mathrm{if\ }j = i, \\
1 & \mathrm{if\ }z_i\ \mathrm{and\ } z_j\ \mathrm{are\ from\ same\ category,}\\ & \mathrm{and\ both\ } q_i\ \mathrm{and\ }q_j>T_{push},\\
0 & \mathrm{otherwise}, 
\end{array} \right.
\end{equation}
where $i$ and $j$ are indices of z for augmented sample. $q_i$ denotes the pseudo label confidence score of the $i$th augmented views. Although class-aware contrastive is theoretically functional, directly using  $T_{push}$ to judge the cleanness of data is problematic since images just above the $T_{push}$ might be out-of-distribution data too. To solve this issue, we propose a foreground re-weighting module to reduce the noise effect on out-of-distribution data. 

\textbf{Foreground Re-weighting} is used to further emphasize training on in-distribution data with high confidence and weaken the potential bias brought from out-of-distribution data. We formulate the final target matrix $W_{target}$ by re-weight $W_{clacon}$ with the foreground probability from model's prediction on $Aug_w$. Simply by multiplying confidence $q$ of two samples, we get a re-weighting factor $w_{re} = q_i \cdot q_j$, where $i$ and  $j$ represent the indices of row and column in $W_{target}$, respectively. Each element $w_{ij}^{target}$ of $W_{target}$ is defined as follows:
\begin{equation}
    w_{ij}^{target} = \left\{ \begin{array}{ll}
q_i\cdot q_j \cdot w_{ij}^{clacon} & \mathrm{if}\ i\ \neq j,\\
 w_{ij}^{clacon} & \mathrm{otherwise}. 
\end{array} \right. \label{eq:target}
\end{equation}

The loss of the class-aware contrastive module $L_c$ is formulated by minimizing the cross-entropy between $S$ in \cref{eq:dotproduct} and $W_{target}$. The loss of $\mathcal{L}_{c}$ can be defined as:

\begin{equation}
    \mathcal{L}_c= H(S, W_{target}) = \sum_{i=1}^{2N}\frac{1}{1+|P(i)|} \mathcal{L}_{c,i}, 
\end{equation}
where $H$ denotes cross entropy and $P(i)$ represents indices of the views from other images of the same class with confidence $p > T_{push}$. $|P(i)|$ denotes its cardinality and $|P(i)| + 1$ represents all positive pairs. $\mathcal{L}_{c,i}$ as below:

\begin{equation}
    \begin{aligned}
\mathcal{L}_{c,i} = &-\mathrm{log}\frac{\mathrm{exp}(z_i\cdot z_i^{*}/\tau) }{\sum_{j=1}^{2N}\mathbbm{1}_{j\ne{i}} \mathrm{exp}(z_i\cdot z_{j}/\tau)}\\
&-\sum_{p\in P(i) }w_{ip}\cdot \mathrm{log}\frac{\mathrm{exp}(z_i\cdot z_p/\tau) }{\sum_{j=1}^{2N}\mathbbm{1}_{j\ne{i}} \mathrm{exp}(z_i\cdot z_{j}/\tau)}, \\
\end{aligned}
\end{equation}

where $z_i^*$ is the embedding of $Aug_s$ rather than $z_i$ from the same image. We calcualte the final total loss \cref{eq:finalloss} using a weighted sum of supervised loss $\mathcal{L}_x$, semi-supervised loss $\mathcal{L}_u$ and class-aware contrastive loss $\mathcal{L}_c$. $\mathcal{L}_x$ and $\mathcal{L}_u$ are from \cref{method:semi-sup-module}. $\lambda_{u}$ and $\lambda_{c}$ are the weights for semi-supervised loss and class-aware contrastive loss, respectively.
\begin{equation}
    \begin{aligned}
    \mathcal{L} = \mathcal{L}_x + \lambda_{u}  \mathcal{L}_u + \lambda_{c}  \mathcal{L}_{c}
    \end{aligned} \label{eq:finalloss}
\end{equation}

\subsection{Applications}

CCSSL method can be used with any pseudo-label-based SSL to alleviate confirmation bias in the training process. Applying CCSSL to SSL methods is simple. We applied the CCSSL to the mainstream SSL methods, like MixMatch\cite{berthelot2019mixmatch}, FixMatch\cite{sohn2020fixmatch} and CoMatch\cite{li2021comatch}. MixMatch and FixMatch are based on pseudo-labels and have a strong augmentation $Aug_s$ for unlabeled data learning. We create another $Aug_s$ for class-aware contrastive module, which trains parallelly with the original network. CoMatch also incorporates contrastive learning, but no class-wise clustering is involved. Since CoMatch already has two $Aug_s$, we directly use them for class-aware contrastive module.  Our experiments in \cref{sec:experiments} show CCSSL not only improves the performance for SSL methods but also speeds up the training process.

\begin{table*}[t]
\centering
\small

\begin{tabular}{c|ccc|ccc|c}
\toprule
    \multirow{2}{*}{Method} &
    \multicolumn{3}{c|}{CIFAR100} & 
    \multicolumn{3}{c|}{CIFAR10} & STL10 \\
\cline{2-8}
 & 400 & 2500 & 10000 & 40 & 250 & 4000 &  \\
\midrule
 Mixmatch~\cite{berthelot2019mixmatch} & 32.39±1.32 & 60.06±0.37 & 71.69±0.33 & 52.46±11.5 & 88.95±0.86 & 93.58±0.10 & 38.02±8.29 \\
 ReMixMatch~\cite{berthelot2019remixmatch}& 55.72±2.06 & 72.57±0.31 & 76.97±0.56 & 80.90±9.64 & 94.56±0.05 & 95.28±0.13 & - \\
 SSWPL~\cite{taherkhani2021self} & - & 73.48±0.45 & 79.12±0.85 & - & - & - & - \\
%  SemCo & / & 68.07±0.01 & 75.55±0.23 & / & 94.88±0.27 & \textbf{96.2±0.08} & / \\
 LaplaceNet~\cite{sellars2021laplacenet} & - & 68.36±0.02 & 73.40±0.23 & - & - & 95.35±0.07 & - \\
 CoMatch~\cite{li2021comatch} & 58.11±2.34 & 71.63±0.35 & 79.14±0.36 & \textbf{93.09±1.39} & \textbf{95.09±0.33} & 95.44±0.20 & 79.80±0.38 \\
 FixMatch~\cite{li2021comatch} & 51.15±1.75 & 71.71±0.11 & 77.40±0.12 & 86.19±3.37 & 94.93±0.65 & \textbf{95.74±0.05} & 65.38±0.42 \\
 \hline
 \textbf{CCSSL(FixMatch)} & \textbf{61.19±1.65} & \textbf{75.7±0.63} & \textbf{80.68±0.16} & 90.83±2.78 & 94.86±0.55 & 95.54±0.20 & \textbf{80.01±1.39} \\
\bottomrule
\end{tabular}
\caption{Top-1 Accuracy for in-distribution datasets including CIFAR100, CIFAR10, and STL10. On high noise-level datasets CIFAR100 and STL10, we achieve the best performance by simply adding CCSSL to Fixmatch. On the easier dataset CIFAR10 with less noise, CCSSL only provides marginal performance gain. '-' means not self-implemented.}
\label{tab:in-distribution-data}
\end{table*}

\section{Experiments} \label{sec:experiments}
In this section, we first evaluate the effectiveness of CCSSL on in-distribution datasets, which may also contain blur out-of-distribution cases. Then, we show CCSSL's ability of confirmations bias alleviation on real-world data. Last, a qualitative comparison is presented. 

\begin{table}
\centering
\begin{tabular}{c|cc|cc}
\toprule
    \multirow{3}{*}{Method} &
    \multicolumn{4}{c}{Semi-iNat 2021} \\
    % \cline{2-5}
    & \multicolumn{2}{c|}{From Scratch} & 
    \multicolumn{2}{C{2cm}}{From MoCo Pretrain} \\
    % & \multicolumn{2}{c|}{Scratch} & 
    % \multicolumn{2}{c}{MoCo Pretrain} \\
\cline{2-5}
& Top1 & Top5 & Top1 & Top5 \\
\midrule
Supervised & 19.09 & 35.85 & 34.96 & 57.11 \\
% MocoV2~\cite{he2020momentum} & - & - & 34.12 & 58.54 \\
\hline
MixMatch~\cite{berthelot2019mixmatch} & 16.89 & 30.83 & - & - \\
\hspace{0.5cm}\textbf{+CCSSL} & \textbf{19.65} & \textbf{35.09} & -  & - \\
\hline
FixMatch~\cite{sohn2020fixmatch} & 21.41 & 37.65 & 40.3 & 60.05 \\
\hspace{0.5cm}\textbf{+CCSSL} & \textbf{31.21} & \textbf{52.25} & \textbf{41.28} & \textbf{64.3} \\
\hline
CoMatch~\cite{li2021comatch} & 20.94 & 38.96 & 38.94 & 61.85 \\
\hspace{0.5cm}\textbf{+CCSSL} & \textbf{24.12} & \textbf{43.23} & \textbf{39.85} & \textbf{63.68} \\
\bottomrule
\end{tabular}
\caption{Accuracy for real-world dataset - Semi-iNat 2021. We evaluated two scenarios: random starting with serious noise and pre-trained starting with less noise. In the noisy random starting scenario, CCSSL achieves significant performance gain by noise alleviation.  In the noiseless pretrained starting scenario, CCSSL achieves marginal improvement because noise has been reduced by the pretrained model in the beginning.}
\label{tab:out-class-data}
\end{table}

\subsection{Datasets}
We assume that the harder the dataset, the higher noise it has, so we conduct experiments on the following different noise level datasets.

\textbf{CIFAR10}~\cite{krizhevsky2009learning} is a relatively easy-level in-distribution dataset consisting of 60K 32x32 colour images in 10 classes, with 6K images per class. 
% \textbf{CIFAR10}~\cite{krizhevsky2009learning} is a relatively easy-level in-distribution dataset comprises 10 classes, with 6K images per class. 
There are 50K training images and 10K test images. We conduct three different experiments on 40, 250, 4000 random selected images in the class balanced way, following~\cite{sohn2020fixmatch}, as in \cref{tab:in-distribution-data}.  

\textbf{STL10}~\cite{coates2011analysis} is a relatively medium-level in-distribution dataset comprises 10 classes. The training set has 5K images (10 predefined folds), 100k unlabeled images containing similar distribution with training images and 800 test images per class. All images are 96x96 pixels. We conduct experiments on all folds following~\cite{sohn2020fixmatch}, as shown in \cref{tab:in-distribution-data}

\textbf{CIFAR100}~\cite{krizhevsky2009learning} is a relatively hard-level in-distribution dataset comprises 100 classes. The training set contains 50K images, while test contains 10K images. All images have a fixed resolution of 32x32. We conduct three different experiments on 400, 2500, 10000 random selected images in a class-balanced way, following~\cite{sohn2020fixmatch}, as shown in \cref{tab:in-distribution-data}. 

\textbf{Semi-iNat 2021}~\cite{su2021semi_iNat} is a real-world dataset designed to expose some of the challenges encountered in a realistic setting, such as significant class imbalance, domain mismatch between the labeled and unlabeled data, and large unknown classes. The labeled training and val part each has 9721 and 4050 images. The unlabeled training part has 313248 images. Experiments in \cref{tab:out-class-data}

\begin{table}[t]
\centering
\small
\setlength\tabcolsep{2pt}

\begin{tabular}{p{0.18\columnwidth}<{\centering}p{0.18\columnwidth}<{\centering}p{0.18\columnwidth}<{\centering}p{0.18\columnwidth}<{\centering}p{0.18\columnwidth}<{\centering}}

\toprule
    {Origin} & {FixMatch} & {CCSSL} & {CoMatch} &{CCSSL}\\
    & & (FixMatch) & & (CoMatch) \\
\midrule
    \centering
    \begin{minipage}[b]{0.18\columnwidth}
		\centering
		\raisebox{-.5\height}{\includegraphics[width=1.0\columnwidth]{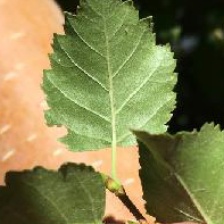}}
	\end{minipage} &
	 \begin{minipage}[b]{0.18\columnwidth}
		\centering
		\raisebox{-.5\height}{\includegraphics[width=1\columnwidth]{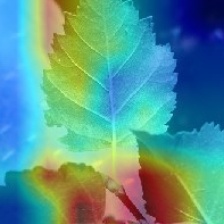}}
	\end{minipage} &
	 \begin{minipage}[b]{0.18\columnwidth}
		\centering
		\raisebox{-.5\height}{\includegraphics[width=1\columnwidth]{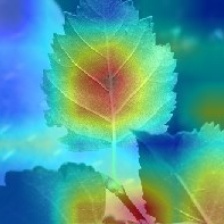}}
	\end{minipage} &
	 \begin{minipage}[b]{0.18\columnwidth}
		\centering
		\raisebox{-.5\height}{\includegraphics[width=1\columnwidth]{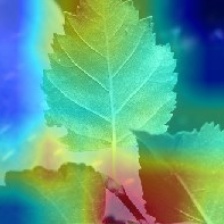}}
	\end{minipage} &
	\begin{minipage}[b]{0.18\columnwidth}
		\centering
		\raisebox{-.5\height}{\includegraphics[width=1\columnwidth]{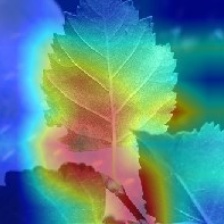}}
	\end{minipage} \\
    \begin{minipage}[b]{0.18\columnwidth}
		\centering
		\raisebox{-.5\height}{\includegraphics[width=1\columnwidth]{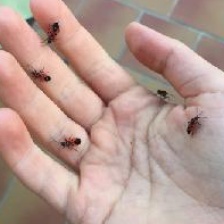}}
	\end{minipage} &
	 \begin{minipage}[b]{0.18\columnwidth}
		\centering
		\raisebox{-.5\height}{\includegraphics[width=1\columnwidth]{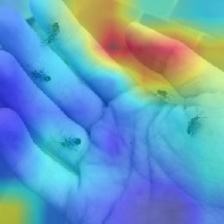}}
	\end{minipage} &
	 \begin{minipage}[b]{0.18\columnwidth}
		\centering
		\raisebox{-.5\height}{\includegraphics[width=1\columnwidth]{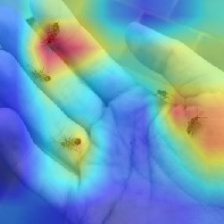}}
	\end{minipage} &
	 \begin{minipage}[b]{0.18\columnwidth}
		\centering
		\raisebox{-.5\height}{\includegraphics[width=1\columnwidth]{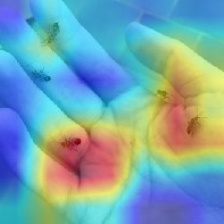}}
	\end{minipage} &
	\begin{minipage}[b]{0.18\columnwidth}
		\centering
		\raisebox{-.5\height}{\includegraphics[width=1\columnwidth]{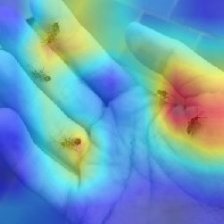}}
	\end{minipage} \\
	
\bottomrule
\end{tabular}

\caption{Qualitative comparison on FixMatch and Comatch with/without CCSSL by Grad-CAM\cite{Grad-CAM}. With CCSSL's de-noise effect, an SSL method focuses more on the foreground objects. CCSSL methods find all ants for the ant image and capture the correct position on the leaf image.}
\label{tab:visulation}
\end{table}

\subsection{In-distribution Data Evaluation}
For CIFAR10, we set $\lambda_{c}=0.2$ with $T_{push}=0$ due to low noise level of the dataset and no out-of-distribution data. For high noise level STL10 and CIFAR100, we set $\lambda_{c}=1.0$ with $T_{push}=0$. We train 512 epochs for all experiments because of fast convergence speed. For other hyperparameters, optimizer and learning rate schedule, our setting is the same as FixMatch~\cite{sohn2020fixmatch}. We compare CCSSL with  MixMatch~\cite{berthelot2019mixmatch}, ReMixMatch~\cite{berthelot2019remixmatch}, FixMatch~\cite{sohn2020fixmatch}, SSWPL~\cite{taherkhani2021self}, LaplaceNet~\cite{sellars2021laplacenet} and Comatch~\cite{li2021comatch}. Since the last three SSL methods can be viewed as an extension of FixMatch, we directly apply CCSSL on FixMatch for a fair comparison. 

In \cref{tab:in-distribution-data}, We find that the performance of CCSSL is higher when the task is noisy and more confirmation bias involved. On CIFAR100 and STL10, the noisiest in-distribution datasets we use, we achieve the best performance only by adding CCSSL on FixMatch. CCSSL is also on par with state-of-the-art methods on CIFAR10. On CIFAR100, we achieve +10.04\%/ +3.99\%/ +3.26\% over origin FixMatch and +3.08\%/ +2.22\%/ +1.54\% over the best of all compared SSL methods for 400, 2500, and 10000 labels respectively. The performance gain decline can be explained by the confirmation bias level decrease with more training data. For the medium hard STL10, we achieve +14.62\% over origin FixMatch and only +0.2\% over the best of all compared SSL methods. CCSSL's performance improvement decreases on the easier dataset with fewer categories. On CIFAR10, we achieve +4.64/ -0.07\%/ -0.2\% over FixMatch and -2.26\%/ -0.23\%/ -0.2\% over the best of all compared SSL methods. With the decrease of noise level, our CCSSL's performance improvement pales compared with other variations of SSL methods tuned for in-distribution datasets but is still helpful for FixMatch. Another thing that needs to be noticed is that when the noise level is the least (CIFAR10 with 4000 labels), FixMatch achieves the best performance by the simplest structure - which means that when the data is clear and enough, the structure of SSL becomes less critical. 

\subsection{Realistic Evaluation} \label{exp:real}
On Semi-iNat 2021, we simulate two situations with different noise levels - (high: training from scratch, low: training from pre-trained model), as shown in \cref{tab:out-class-data}. The results show that CCSSL can improve SSL in both situations but more noise more improvement. 

For Semi-iNat 2021, we use an image size of 224x224, resnet50 backbone, projection head with a dimension of 64 like in~\cite{he2020momentum}, and training on 4 V100 with batchsize 64 for labeled data and batchsize 448 with $\mu=7$ for unlabeled data. For FixMatch~\cite{sohn2020fixmatch}, pseudo label thresh $\tau=0.8$ achieves the best results, but 0.6 when combining with our method. For MixMatch~\cite{berthelot2019mixmatch} and CoMatch~\cite{li2021comatch}, we directly follow their papers' setting. We set  $\lambda_{c}=2.0$ with $T_{push}=0.9$ due to high noise level on out-of-distribution data. When training from Self-SL pre-trained models, we freeze the first three blocks of resnet50 because of the forgetting problem. Other settings are the same as~\cite{su2021realistic}. 
% For experiments, we first verify the performance change of MixMatch, FixMatch, and CoMatch with and without of CCSSL. Then, we evaluate from a MoCo\cite{he2020momentum} pre-trained model, which makes SSL methods start with less noise. 

In \cref{tab:out-class-data} training from scratch, state-of-the-art SSL methods only have on-par performance with the supervised baseline. Compared to the performance on in-distribution datasets, the noise in pseudo-labels harms SSL methods by a large margin. With class-aware contrastive module, we are able to alleviate confirmation bias on out-of-distribution data while maintaining downstream tasks' clustering ability. We improve state-of-the-art SSL by a large margin when using with CCSSL (+2.7\%/ +9.8\%/ +3.1\% on MixMatch, FixMatch and CoMatch separately). The experiments show that CCSSL can be a big helper for pseudo-label-based SSL to reduce the negative impact from noisy predictions when training from scratch on the real-world dataset.

In \cref{tab:out-class-data}, training from MoCo Semi-iNat pre-trained model simulates the situation where SSL starts with less noise~\cite{caron2021emerging}. CCSSL only improves +1.05\%/ +0.91\%/ on FixMatch and CoMatch. The result shows that when a model starts with less noise, our proposed CCSSL can also reduce noise introduced in the training process. 

\subsection{Qualitative Evaluation}
In \cref{tab:visulation}, we show the qualitative results on FixMatch and CoMatch with/without CCSSL. With CCSSL's de-noise effect, an SSL method focuses more on the foreground objects. CCSSL methods find all ants for the ant image and capture the correct position on the leaf image.  We choose FixMatch and CoMatch for visualization because the two are the best on Semi-iNat among SSL methods we tested.

\begin{figure}[t]
  \centering
  \includegraphics[width=1.0\linewidth]{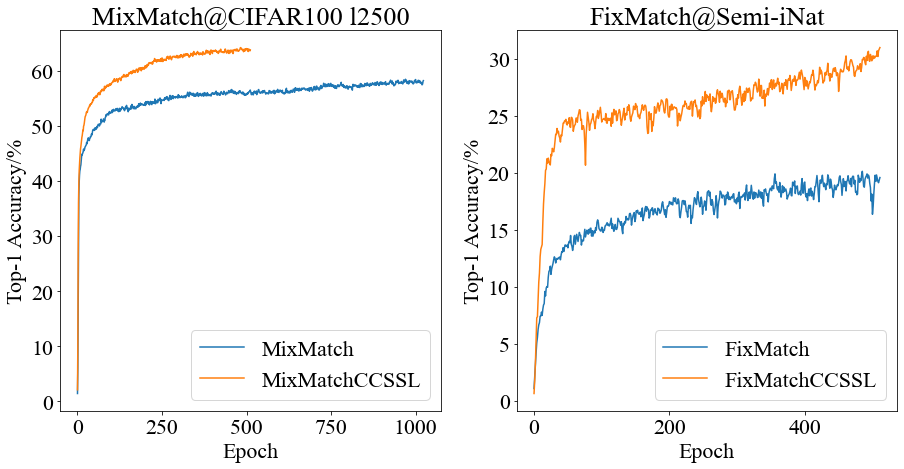}
  \caption{Convergence speed and performance difference on MixMatch and FixMatch with/without CCSSL. As shown in the figure, CCSSL methods achieves better performance with much fewer training epochs.}
  \label{fig:worwithout_ccssl}
\end{figure}

\subsection{Convergence Speed}
In \cref{fig:worwithout_ccssl}, we study the convergence speed using CCSSL on MixMatch and FixMatch. We observe that CCSSL methods achieve the best performance with much fewer training iterations. This phenomenon is the same when training on clean pseudo-labels~\cite{li2021comatch} because of the noise alleviation effect of CCSSL. More specifically, better pseudo labels in the early phase helps better guide the learning process and translate to faster convergence.
% As in \cref{tab:pseudo-label}, the model's pseudo label quality is better with CCSSL on many data settings. The more noise (less labels), the more salient of improvements.

\section{Ablation}
\subsection{Contrastive \& Class-Aware \& Re-weighting}
\begin{table}
\centering
\small
\begin{tabular}{C{1cm}|C{1.5cm}|C{1cm}|C{1.3cm}|C{1.3cm}}
\toprule
contrastive & re-weighting & class-aware & Semi-iNat & CIFAR100\\
\midrule
 &  &  & 21.41 & 72.54\\
\checkmark &  &  & 27.73 & 72.37\\
\checkmark & \checkmark &  & 29.48 & 72.67\\
\checkmark &  & \checkmark & 30.49 & 75.89 \\
\checkmark & \checkmark & \checkmark & \textbf{31.21} & \textbf{75.96} \\
\bottomrule
\end{tabular}
\caption{Accuracy for different combinations of contrastive learning, class-aware and re-weighting.}
\label{tab:ablation-for-module}
\end{table}

% \begin{table}
% \centering
% \small
% \begin{tabular}{c|m{0.03\textwidth}m{0.03\textwidth}m{0.04\textwidth}|m{0.03\textwidth}m{0.03\textwidth}m{0.035\textwidth}}
% \toprule
%     % \multirow{2}{*}{Best pseudo accuracy} &
%     Best Pseudo &
%     \multicolumn{3}{c|}{CIFAR100} & 
%     \multicolumn{3}{c}{CIFAR10} \\
% \cline{2-7}
% Accuracy & 400 & 2500 & 10000 & 40 & 250 & 4000 \\
% \midrule
% FixMatch & 62.45 & 83.82 & 92.01 & 94.81 & 96.09 & \textbf{97.79} \\
% \hspace{0.5cm}\textbf{+CCSSL} & \textbf{69.79} & \textbf{85.46} & \textbf{92.08} & \textbf{96.01} & \textbf{96.51} & 97.59 \\
% \bottomrule
% \end{tabular}
% \caption{Best pseudo-label accuracy for various data setttings.}
% \label{tab:pseudo-label}
% \end{table}

In ~\cref{tab:ablation-for-module}, we investigate three main techniques of CCSSL, including contrastive learning, class-aware, and re-weighting, based on FixMatch. We observe that all components are helpful. However, directly using contrastive learning deteriorates the performance on in-distribution dataset CIFAR100 while largely benefiting the real-world dataset Semi-iNat 2021 with extensive out-of-distribution data. CCSSL found the equilibrium point for contrastive and clustering by class-aware contrastive and re-weighting, and improved FixMatch by a large margin on both in-distribution and real-world datasets.

\subsection{Ratio of Unlabeled Data}

\begin{table}
\centering
\small
\begin{tabular}{c|c|c}
\toprule
Ratio & \multicolumn{2}{c}{Semi-iNat}  \\
($\mu$) & $T_{push}=0$ & $T_{push}=0.9$ \\
\midrule
2 & 23.01 & 21.93 \\
4 & 23.88 & 25.19 \\
5 & \textbf{25.33} & 26.81 \\
6 & 22.32 & 24.47 \\
7 & 23.14 & \textbf{27.46} \\
\bottomrule
\end{tabular}
\caption{Different ratios of unlabeled data on Semi-iNat with different degrees of class-aware ($T_{push}$). When $T_{push} = 0$ (no contrastive involved), smaller ratios achieves better results because of noise in unlabeled data. After setting $T_{push} = 0.9$ to regularize noise, larger ratio achieves better results.}
\label{tab:ablation-for-mu}
\end{table}
As shown in \cref{tab:ablation-for-mu}, we experiment on different values of $\mu$, which defines the ratio of labeled and unlabeled data in a batch. We observe that when only using class-aware clustering without contrastive regularization ($T_{push}=0$),  CCSSL achieves the best result with a small ratio of unlabeled data ($\mu=5$). However, after we increase the effect of contrastive learning by changing $T_{push}=0.9$ (sample conficence $<$ 0.9 will do contrastive learning), the larger ratio has better performance ($\mu=7$). Less noisy dataset like CIFAR100 also benefits from large unlabeled data ratio~\cite{nassar2021labels}. This phenomenon can be explained by the higher noise amount introduced from self-generated labels with larger ratio of unlabeled data. With CCSSL's noise alleviation, SSL methods achieve the best performance with a larger unlabeled ratio.
% Without noise reduction on low confidence out-of-distribution unlabeled data, the larger the ratio, the larger the confirmation bias. 

\subsection{Threshold for Class-Aware and Contrastive}
\begin{table}
\centering
\small
\begin{tabular}{c|c|c}
\toprule
$T_{push}$ & Semi-iNat & CIFAR100@l2500 \\
\midrule
0 & 23.88 & 75.59 \\
0.4 & 23.38 & \textbf{75.96} \\
0.6 & 24.49 & 75.76 \\
0.7 & 24.64 & 75.9 \\
0.8 & 23.63 & 75.7 \\
0.9 & \textbf{25.21} & 75.79 \\
\bottomrule
\end{tabular}
\caption{Experiments on different class-aware thresholds. Our algorithm is robust with varying $T_{push}$  from 0.6 to 0.9. Semi-iNat has a larger noise level and needs larger $T_{push}$ which proves CCSSL's ability for noise alleviation.}
\label{tab:ablation-for-push}
\end{table}

We found that our algorithm is robust with varying the threshold $T_{push}$ from 0.6 to 0.9. However, the optimal threshold for class-aware and contrastive learning is inconsistent on different noise level datasets. For high noise level Semi-iNat with unbalanced distribution and unknown classes, $T_{push} = 0.9$ achieves the best performance. For low noise level CIFAR100, $T_{push} = 0.4$ achieves the best performance. The result proves CCSSL's noise-reducing ability by needing to regularize more on noisy datasets. 

\section{Limitations}
Potential harmful training fluctuation does exist in our experiments: performance on Semi-iNat fluctuates more than on CIFAR100 in the \cref{fig:worwithout_ccssl}. The main reason lies in the lower quality of pseudo-labels for high noise-level data. Increasing the weight for the class-aware  contrastive module or training with different seeds to average the effect of fluctuation can alleviate the problem. During our experiments, we fixed the weight and seed for a fair comparison. 

\section{Conclusion}
We proposed CCSSL, a general confirmation bias alleviation method for pseudo-label-based SSL methods. CCSSL reduces noise by constructing another feature space instead of using a model's output. By class-aware contrastive module, we apply image-level contrastive on out-of-distribution data for noise alleviation and class-level clustering on in-distribution data for blending into downstream tasks.  We have conducted extensive experiments over different noise level datasets, including in-distribution datasets and a real-world dataset,  to demonstrate the effectiveness of CCSSL. With simply replacing the semi-supervised module, we improve the state-of-the-art SSL~\cite{sohn2020fixmatch}~\cite{berthelot2019mixmatch}~\cite{li2021comatch} by a large margin.
With the effort above, CCSSL is proved to be helpful in making SSL more practical in the real world. 
% With the effort above, we proved that CCSSL makes a big step forward to make SSL more practical in the real world.

\textbf{Potential negative societal impact} CCSSL is a fundamental technology that can generate robust semi-supervised classification models in the real world. Maliciously using a classification model to negatively impact society is possible. 

%%%%%%%%% REFERENCES
{\small
\bibliographystyle{ieee_fullname}
\bibliography{egbib}
}

\end{document}

\title{`Class-Aware Contrastive Semi-Supervised Learning' Supplementary for FAQs}  % **** Enter the paper title here

\maketitle
\thispagestyle{empty}
\appendix

%%%%%%%%% BODY TEXT - ENTER YOUR RESPONSE BELOW
\begin{table}
\centering
\small
\begin{tabular}{c|m{0.03\textwidth}m{0.03\textwidth}m{0.04\textwidth}|m{0.03\textwidth}m{0.03\textwidth}m{0.035\textwidth}}
\toprule
    % \multirow{2}{*}{Best pseudo accuracy} &
    Best Pseudo &
    \multicolumn{3}{c|}{CIFAR100} & 
    \multicolumn{3}{c}{CIFAR10} \\
\cline{2-7}
Accuracy & 400 & 2500 & 10000 & 40 & 250 & 4000 \\
\midrule
FixMatch & 62.45 & 83.82 & 92.01 & 94.81 & 96.09 & \textbf{97.79} \\
\hspace{0.5cm}\textbf{+CCSSL} & \textbf{69.79} & \textbf{85.46} & \textbf{92.08} & \textbf{96.01} & \textbf{96.51} & 97.59 \\
\bottomrule
\end{tabular}
\caption{Best pseudo-label accuracy for various data setttings.}
\label{tab:pseudo-label}
\end{table}

\begin{figure}[t]
  \centering
  \includegraphics[width=1.0\linewidth]{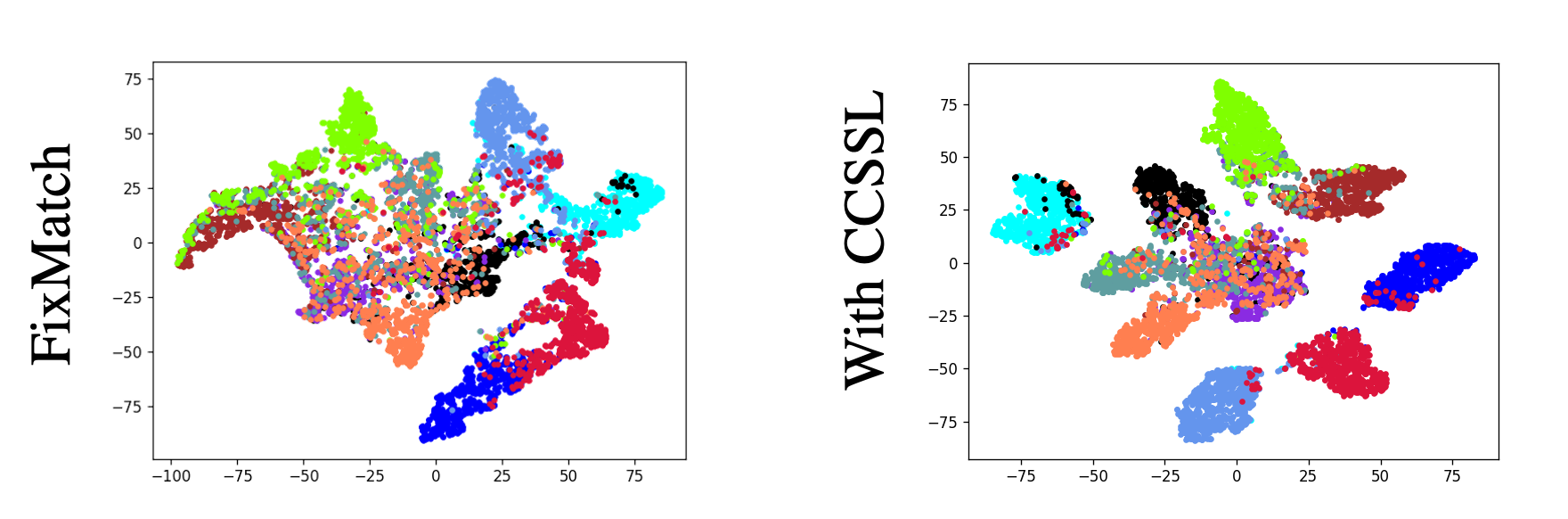}
   \caption{TSNE of high dimensional features with/without CCSSL. }
   \label{fig:tsne}
\end{figure}

\begin{figure}[t]
  \centering
  \includegraphics[width=1.0\linewidth]{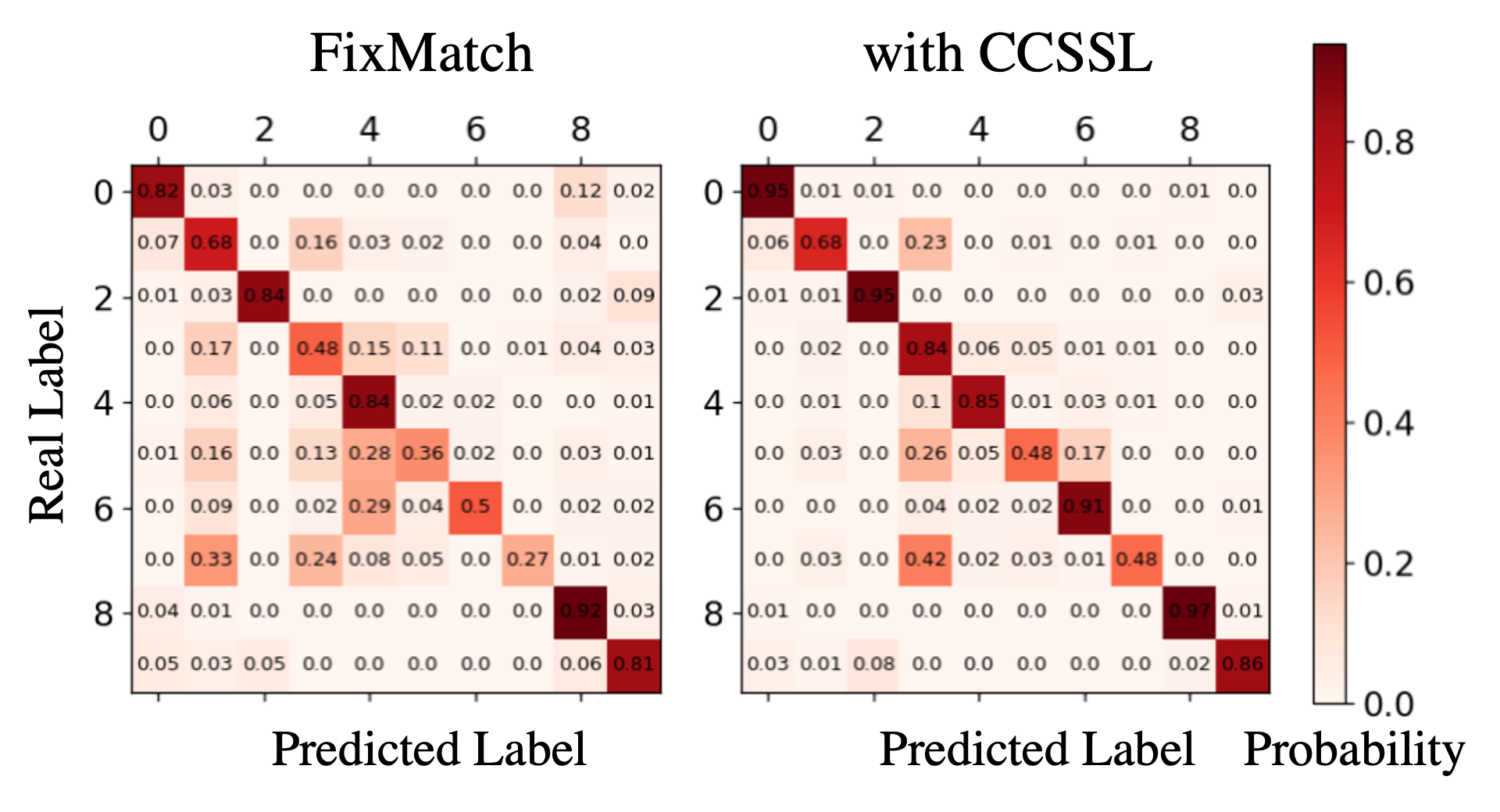}
   \caption{Confusion Matrix with/without CCSSL. STL10 is used for both because CIFAR10 is easy and CIFAR100 has too many categories to be visualized.}
   \label{fig:cf_matrix}
\end{figure}

\section{What is the intuition of using contrastive in SSL? Are there some proofs other than the final performance?}

The intuition of CCSSL is to end-to-end reduce noise on the feature level by contrastive learning while maintaining model's clustering ability.

\textbf{(1) Noise reduction by a projection head.} The noise reduction is on the feature level and not directly on pseudo labels. \textbf{Qualitatively}, ~\cite{wang2021understanding} verified that the uniformity of embeddings from the projection head is essential to learn general separable features. As in ~\cref{fig:tsne}, the features are more separable with CCSSL. \textbf{Quantitatively}, the model's confusion on similar categories is halved with CCSSL, as in ~\cref{fig:cf_matrix}. With CCSSL, the model can be more resilient to misclassification and thus has less noise on pseudo-labels. So if the model starts with a good pretrained model with good features, like MoCo, CCSSL can only benefit the training process and less helpful. 

% pseudo-label quality, but there are no quantitative results to support this other than the final performance.
\textbf{(2) Quantitative analysis for pseudo-labels.} As in \cref{tab:pseudo-label}, the model's pseudo label quality is better with CCSSL on many data settings. The more noise (less labels), the more salient of improvements. We cannot evaluate pseudo-label accuracy on Semi-iNat 2021 and STL10 because labels are not provided. We will provide more evaluation results on our public github page https://github.com/TencentYoutuResearch/Classification-SemiCLS.

% \section{What is the relationship with supervised contrastive learning \cite{khosla2020supervised}?}

% \textbf{(1) Novelty}. To our best knowledge, our work is the first to deal with noisy out-of-distribution (OOD) data with contrastive learning for semi-supervised learning.

% \textbf{(2) Originality}. The class-aware contrastive module is formulated by in-distribution data clustering and out-of-distribution data contrasting with re-weighting. \cite{khosla2020supervised} is a non re-weighting version of in-distribution data clustering (class-aware) and can only be used with ground-truth labels. In short, \cite{khosla2020supervised} is a special version of CCSSL and cannot be directly applied to SSL because of noise contained in pseudo-labels. Also in ablation Tab. 4, \cite{khosla2020supervised} used in CCSSL only provides +2.75\% from overall +9.8\% improvement.

\section{Details of method formulation}

(1) \textbf{Relationship with infoNCE.} Our method improves upon the \textbf{infoNCE}, so we provide $S$, $W_{con}$ and $\mathcal{L}_\text{infoNCE}$ as references and has not directly used them in the final equation for simplicity.

(2) \textbf{Cross Entropy of $\mathcal{L}_c$.} The \textbf{cross-entropy} between $S$ and $W_{target}$ for $\mathcal{L}_c$ in Eq. (9) means that we take columns as classes and rows as samples to calculate the soft cross-entropy between two matrices. Soft means the targets have been re-weighted. 

(3) \textbf{Meaning of $\mathbf{P_i}$.} $\mathbf{z_p}$ is the high dimensional feature of sample $p$ from $\mathbf{P_i}$, which represents the sample set with the same pseudo-label of sample $i$.

\section{Details of training}

The training strategy is generally the same as FixMatch for a fair comparison. 

(1) \textbf{Warm-up Peroid}. There is no warm-up period, and CCSSL can help a SSL model from a noisy starting. Because the confidence is low in the beginning, the contrastive part of CCSSL is triggered and helps the model learn a general representation instead of over-fitting on the training data. After the model learns a few epochs and can differentiate in-distribution and out-of-distribution data, the clustering part of CCSSL will then be triggered with less noise.

(2) \textbf{Augmentations}. The augmentations of supervised and semi-supervised branches are the same as FixMatch, and the strong augmentation of CCSSL branch is the same as MoCo for a fair comparison. 

(3) \textbf{Backbone}. The model size is the same as FixMatch during inference and the backbone follows the convention with wid-resnet 28-2, 28-8 for CIFAR10 and CIFAR100, resnet18 for STL10,  resnet50 for Semi-iNat.

%%%%%%%%% REFERENCES
{\small
\bibliographystyle{ieee_fullname}
\bibliography{egbib}
}